\documentclass[runningheads]{llncs}
\usepackage{graphicx}
\begin{document}
\title{L3Cube-MahaNER: A Marathi Named Entity Recognition Dataset and BERT models\thanks{Supported by L3Cube Pune}}
\titlerunning{L3Cube-MahaNER: A Marathi Named Entity Recognition Dataset}
%
\author{Parth Patil\inst{1,3} \and
Aparna Ranade\inst{1,3} \and
Maithili Sabane\inst{1,3} \and
Onkar Litake\inst{1,3} \and
Raviraj Joshi\inst{2,3}
}
\authorrunning{O. Litake et al.}
%
\institute{Pune Institute of Computer Technology, Pune, Maharashtra India \and
Indian Institute of Technology Madras, Chennai, Tamilnadu India \and
L3Cube Pune \\
\email{\{parthpatil8399,aparna.ar217,msabane12\}@gmail.com, }\email{onkarlitake@ieee.org, }\\
\email{ravirajoshi@gmail.com}}
\maketitle              
\begin{abstract}
Named Entity Recognition (NER) is a basic NLP task and finds major applications in conversational and search systems. It helps us identify key entities in a sentence used for the downstream application. NER or similar slot filling systems for popular languages have been heavily used in commercial applications. In this work, we focus on Marathi, an Indian language, spoken prominently by the people of Maharashtra state. Marathi is a low resource language and still lacks useful NER resources. 
We present L3Cube-MahaNER, the first major gold standard named entity recognition dataset in Marathi. We also describe the manual annotation guidelines followed during the process. In the end, we benchmark the dataset on different CNN, LSTM, and Transformer based models like mBERT, XLM-RoBERTa, IndicBERT, MahaBERT, etc. The MahaBERT provides the best performance among all the models.  
The data and models are available at {\footnotesize https://github.com/l3cube-pune/MarathiNLP}\\

\keywords{Named Entity Recognition \and NER \and Marathi Dataset \and Transformers}
\end{abstract}

\section{Introduction}

A principal technique of information extraction is Named Entity Recognition. It is an integral part of natural language processing systems. The technique involves the identification and categorization of the named entity \cite{marrero2013named,lample2016neural}. These categories include entities like people's names, locations, numerical values, and temporal values. NER has a myriad of applications like customer service, text summarization, etc.
Through the years, a large amount of work has been done for Named Entity Recognition in the English language \cite{yadav2018survey}. The work is very mature and the functionality comes out of the box with NLP libraries like NLTK \cite{nltk} and spacy \cite{spacy}. In contrast, limited work is done in the Indic languages like Hindi and Marathi \cite{kale2017survey}. \cite{Patil:16} addresses the problems faced by Indian languages like the presence of abbreviations, ambiguities in named entity categories, different dialects, spelling variations and the presence of foreign words. \cite{Shah:16} elaborates on these issues along with others like the lack of well-annotated data, fewer resources, and tools, etc. Furthermore, the existing resources for NER in Marathi released in \cite{murthy-etal-2018-judicious}  titled IIT Bombay Marathi NER Corpus has only 3588 train sentences and 3 target named entities. Also, about 39 percent of sentences in this dataset contain O tags only further reducing the number of useful tokens. Moreover, many datasets aren’t available publicly or contain fewer sample sentences. We aim to build a much bigger Marathi NER corpora with a variety of labels currently missing in the literature. Although, text classification in Hindi and Marathi has recently received some attention \cite{joshi2019deep,kulkarni2022experimental,kulkarni2021l3cubemahasent,velankar2021hate}, however the same is not true for NER. 

In this paper, we present our dataset L3Cube-MahaNER. This dataset has been manually annotated and compiled in-house. It is the largest publicly available dataset for Marathi NER annotated according to the IOB and non-IOB notation.  It contains 25,000 manually tagged sentences categorized according to the eight entity classes. The original sentences have been taken from a news domain corpus \cite{joshi2022l3cube} and the average length of these sentences is 9 words.  These entities annotated in the dataset include names of locations, organizations, people, and numeric quantities like time, measure, and other entities like dates and designations. The paper also describes the dataset statistics and the guidelines that have been followed while tagging these sentences. 

We also present the results of deep-learning models like Convolutional Neural Network (CNN), Long-Short Time Memory (LSTM), biLSTM, and Transformer models like mBERT \cite{devlin2019bert}, IndicBERT \cite{indicbert}, XLM-RoBERTa, RoBERTa-Marathi, MahaBERT \cite{joshi2022l3cube}, MahaROBERTa, MahaALBERT that have been trained on the L3Cube-MahaNER dataset. We experiment on all major multi-lingual and Marathi BERT models to establish a benchmark for future comparisons. 
The MahaBERT model model fined-tuned on L3Cube-MahaNER is termed as MahaNER-BERT\footnote{https://huggingface.co/l3cube-pune/marathi-ner} and is shared publicly on model hub. All the resources are publicly shared on github\footnote{https://github.com/l3cube-pune/MarathiNLP}.


\section{ Related Work}

Named Entity Recognition is a concept that originated at the Message Understanding Conferences \cite{a} in 1995. Machine learning techniques and linguistic techniques were the two major techniques used to perform NER. Handmade rules \cite{b} developed by experienced linguists were used in the linguistic techniques. These systems, which included gazetteers, dictionaries, and lexicalized grammar, demonstrated good accuracy levels in English. However, these strategies had the disadvantage of being difficult to transfer to other languages or professions.  Decision Trees \cite{c}, Conditional Random Field, Maximum Entropy Model \cite{d}, Hidden Markov Model, Support Vector Machine were included in machine learning techniques. To attain better competence, these supervised learning algorithms make use of massive volumes of NE annotated data.

A comparative study by training the models on the same data using  Support Vector Machine (SVM) and Conditional Random Field(CRF) was carried out by \cite{e}. It was concluded that the CRF model was superior. A more effective hybrid system consisting of the Hidden Markov Model, a combination of handmade rules and MaxEnt was introduced by \cite{f} for performing NER. Deep learning models were then utilized to complete the NER problem as technology progressed. CNN \cite{g}, LSTM \cite{h}, biLSTM \cite{i}, and Transformers were among the most popular models.

NER for Indian languages is a comparatively difficult task due to a lack of capitalization, spelling variances, and uncertainty in the meaning of words. The structure of the language is likewise difficult to grasp. Furthermore, the lack of a well-ordered labeled dataset makes advanced approaches such as deep learning methods difficult to deploy. \cite{j} has described various problems faced while implementing NER for Indian languages. 

\cite{murthy-etal-2018-judicious} introduced Marathi annotated dataset named IIT Bombay Marathi NER Corpus for Named Entity Recognition consisting of 5591 sentences and 108359 tags. They considered 3 main categories named Location, Person, Organization for training the character-based model on the dataset. They made use of multilingual learning to jointly train models for multiple languages, which in turn helps in improving the NER performance of one of the languages.

\cite{l} in 2017 released a dataset named WikiAnn NER Corpus consisting of 14,978 sentences and 3 tags labeled namely Organization, Person, and Location. It is however a silver-standard dataset for 282 different languages including Marathi. This project aims to create a cross-lingual name tagging and linking framework for Wikipedia's 282 languages.

\section{Compilation of dataset}

\subsection{Data Collection}
Our dataset consists of 25,000 sentences in the Marathi language. We have used the base sentences from the L3Cube-MahaCorpus \cite{joshi2022l3cube}, which is a monolingual Marathi dataset majorly from the news domain. \\
The sentences in the dataset are in the Marathi language with minimal appearance of English words and numerics as present in the original news. However, while annotating the dataset, these English words have not been considered as a part of the named entity categories. Furthermore, the dataset does not preserve the context of the news, such as the publication profiles, regions, and so on.

\subsection{Dataset Annotation}
We have manually tagged the entire dataset into eight named entity classes. These classes include Person (NEP), Location(NEL), Organization(NEO), Measure(NEM), Time(NETI), Date(NED), and Designation(ED).
While tagging the sentences, we established an annotation guideline to ensure consistency. The first 200 sentences were tagged together to further establish consistency among four annotators proficient in Marathi reading and writing. Post this the tagging was performed in parallel except for ambiguous sentences which were separately handled.
Firstly, the sentences were relieved of any contextual associations. 
Then, the approach for the contents of the named entity classes was decided as follows. Proper nouns involving persons' names are tagged as NEP and places are tagged as NEL. All kinds of organizations like companies, councils, political parties, and government departments are tagged as NEO. Numeric quantities of all kinds are tagged as NEM with respect to the context. Furthermore, temporal values like time are tagged as NETI, and dates are tagged as NED. Apart from that, individual titles and designations, which precede proper nouns in the sentences are tagged as ED.
Despite maintaining these guidelines, some entities had ambiguous meanings and were difficult to tag. In these circumstances, we resolved the intricacies unanimously by taking a vote amongst the annotators. The sentences were tagged according to the predominant vote.  

\subsection{Dataset Statistics}

\begin{table}
\centering
\caption{Count of sentences and tags in the L3Cube-MahaNER}
\begin{tabular}{p{2cm}p{2cm}p{2cm}}
\hline
\textbf{Dataset} & \textbf{Sentence Count} & \textbf{Tag Count}\\
\hline
{Train} & {21500} & {26502} \\
{Test} & {2000} & {2424} \\
{Validation} & {1500} & {1800} \\
\hline
\end{tabular}
\begin{tabular}{lc}
\hline
\end{tabular}
\end{table}

\begin{table}
\centering
\caption{Count of individual tags in L3Cube-MahaNER}
\begin{tabular}{p{2cm}p{2cm}p{2cm}p{2cm}}
\hline
\textbf{Tags} & \textbf{Train} & \textbf{Test} & \textbf{Validation}\\
\hline
{NEM} & {7052} & {620} & {488} \\
{NEP} & {6910} & {611} & {457} \\
{NEL} & {4949} & {447} & {329} \\
{NEO} & {4176} & {385} & {268} \\
{NED} & {2466} & {244} & {182} \\
{ED} & {1003} & {92} & {75} \\
{NETI} & {744} & {73} & {48} \\
\hline
\end{tabular}
\begin{tabular}{lc}
\hline
\end{tabular}
\end{table}

\begin{table}
\centering
\caption{Count of individual tags of L3Cube-MahaNER}
\begin{tabular}{p{2cm}p{2cm}p{2cm}p{2cm}}
\hline
\textbf{Tags} & \textbf{Train} & \textbf{Test} & \textbf{Validation}\\
\hline
{B-NEM} & {5824} & {523} & {404} \\
{I-NEM} & {1228} & {97} & {84} \\
{B-NEP} & {4775} & {428} & {322} \\
{I-NEP} & {2135} & {183} & {135} \\
{B-NEL} & {4461} & {407} & {293} \\
{I-NEL} & {488} & {40} & {36} \\
{B-NEO} & {2741} & {256} & {178} \\
{I-NEO} & {1435} & {129} & {90} \\
{B-NED} & {1937} & {191} & {141} \\
{I-NED} & {529} & {53} & {41} \\
{B-ED} & {838} & {74} & {61} \\
{I-ED} & {165} & {18} & {14} \\
{B-NETI} & {633} & {63} & {43} \\
{I-NETI} & {111} & {10} & {5} \\
\hline
\end{tabular}
\begin{tabular}{lc}
\hline
\end{tabular}
\end{table}

For more clarity, some example sentences with tagged entities are mentioned in Table 6.


\section{Experimental Techniques} 

\subsection{Model Architectures}
The deep learning models are trained using large labeled datasets and the neural network architectures learn features from the data effectively, without the need for feature extraction to be done manually.\\
Similarly, the transformer aims to address sequence-to-sequence problems while also resolving long-range relationships in natural language processing. The transformer model contains a "self-attention" mechanism that examines the relationship between all of the words in a phrase. It provides differential weightings to indicate which phrase components are most significant in determining how a word should be read. Thus the transformer identifies the context that assigns each word in the sentence its meaning. The training time also is lowered as the feature enhances parallelization.
\\ \\
\textbf{CNN:} This model uses a single 1D convolution over the 300-dimensional word embeddings. These embeddings are fed into a Conv1D layer having 512 filters and a filter size of 3. The output at each timestep is subjected to a dense layer of size 8. The dense layer size is equal to the size of the output labels. There are 8 output labels for non-IOB notation and 15 output labels for IOB notation. The activation function used is relu. All the models have the same optimizer and loss functions. The optimizer used is RMSPROP. The embedding layer for all the word-based models is initialized using fast text word embeddings.  
\\ \\
\textbf{LSTM:} This model uses a single LSTM layer to process the 300-dimensional word embeddings. The LSTM layer has 512 hidden units followed by a dense layer similar to the CNN model. 
\\ \\
\textbf{biLSTM:} It is analogous to the CNN model with the single 1D convolution substituted by a biLSTM layer.  An embedding vector of dimension 300 is used in this model and the biLSTM has 512 hidden units. A batch size of 16 is used. 
\\ \\
\textbf{BERT:} BERT \cite{bert} is a Google-developed transformer-based approach for NLP pre-training that was inspired by pre-training contextual representations.
It's a deep bidirectional model, which means it's trained on both sides of a token's context. BERT's most notable feature is that it can be fine-tuned by adding a few output layers.
\\ \\
\textbf{mBERT:} mBERT \cite{mbert}, which stands for multilingual BERT is the next step in constructing models that understand the meaning of words in context. A deep learning model was built on 104 languages by concurrently encoding all of their information on mBERT.
\\ \\
\textbf{ALBERT:} ALBERT \cite{albert} is a transformer design based on BERT that requires many fewer parameters than the current state-of-the-art model BERT. These models can train around 1.7 times quicker than BERT models and have greater data throughput than BERT models. IndicBERT is a multilingual ALBERT model that includes 12 main Indian languages and was trained on large-scale datasets. Many public models, such as mBERT and XLM-R, have more parameters than IndicBERT, although the latter performs exceptionally well on a wide range of tasks.
\\ \\
\textbf{RoBERTa:} RoBERTa \cite{roberta} is also an unsupervised transformers model that has been trained on a huge corpus of English data. This means it was trained exclusively on raw texts, with no human labeling, and then utilized an automated approach to generate labels and inputs from those texts. The multilingual model XLM-RoBERTa has been trained in 100 languages. Unlike certain XLM multilingual models, it does not require lang tensors to detect which language is being used. It can also deduce the correct language from the supplied ids.

%

\begin{table*}
\begin{center}
\caption{\label{result-tab} F1 score(macro), precision and recall of various transformer and normal models for IOB notation using the  Marathi dataset.}
\begin{tabular}{{p{5cm}p{2cm}p{2cm}p{2cm}p{2cm}}}
\hline \textbf{Model} & \textbf{F1} & \textbf{Precision} & \textbf{Recall} & \textbf{Accuracy} \\
\hline

     mBERT & 82.82 & 82.63 & 83.01 & 96.75 \\
    \hline
    Indic BERT & 84.66 & 84.10 & 85.22 & 97.09\\
    \hline
    XLM-RoBERTa & 84.19 & 83.42 & 84.97 & 97.12 \\
    \hline
    RoBERTa-Marathi & 81.93 & 81.58 & 82.29 & 96.67 \\
    \hline
    MahaBERT & 84.81 & 84.55 & 85.07 & 97.10\\
    \hline
    MahaRoBERTa & \textbf{85.30} & \textbf{84.27} &  \textbf{86.36} & \textbf{97.18} \\
    \hline
    MahaAlBERT & 84.50 & 84.54 & 84.45 & 96.98\\
    \hline
    CNN & 72.2 & 81.0 & 66.6 & 97.16 \\
    \hline
    LSTM &  70.0 & 77.1 & 64.8 & 94.46 \\
    \hline
    biLSTM & 73.7 & 77.2 & 77.6 & 94.99 \\
    \hline
    
\end{tabular}
\end{center}
\end{table*}

\begin{table*}
\begin{center}
\caption{\label{result-tab} F1 score(macro), precision and recall of various transformer and normal models for non-IOB notation using the  Marathi dataset.}
\begin{tabular}{{p{5cm}p{2cm}p{2cm}p{2cm}p{2cm}}}
\hline \textbf{Model} & \textbf{F1} & \textbf{Precision} & \textbf{Recall} & \textbf{Accuracy} \\
\hline

     mBERT & 85.3 & 82.83 & 97.94 & 96.92 \\
    \hline
    Indic BERT & 86.56 & 85.86 & 87.27 & 97.15 \\
    \hline
    XLM-RoBERTa & 85.69 & 84.21 & 87.22 & 97.07 \\
    \hline
    RoBERTa-Marathi & 83.86 & 82.22 & 85.57 & 96.92 \\
    \hline
    MahaBERT & \textbf{86.80} & \textbf{84.62} & \textbf{89.09} & \textbf{97.15}\\
    \hline
    MahaRoBERTa & 86.60 & 84.30 & 89.04 & 97.24\\
    \hline
    MahaAlBERT & 85.96 & 84.32 & 87.66 & 97.32\\
    \hline
    CNN & 79.5 & 82.1 & 77.4 & 97.28 \\
    \hline
    LSTM &  74.9 & 84.1 & 68.5 & 94.89 \\
    \hline
    biLSTM & 80.4 & 83.3 & 77.6 & 94.99 \\
    \hline
    
\end{tabular}
\end{center}
\end{table*}

\section{Results}

In this study, we have experimented with various model architectures like CNN, LSTM, biLSTM, and transformers like BERT, RoBERTa to perform named entity recognition on our dataset. This section presents the F1 scores attained by training these models on our dataset for IOB and non-IOB notations. The results have been reported in Table 4 and Table 5 respectively. Among the CNN and LSTM based models, the biLSTM model with the trainable word embeddings gives the best results on the L3Cube-MahaNER dataset for IOB as well as non-IOB notations. Moreover, for the transformers-based models, it is observed that the MahaRoBERTa model yields the best results for IOB and MahaBERT provides the best results for non-IOB notations. The LSTM and the RoBERTa-Marathi models report the lowest scores among all models for both.
\\

\begin{table}
\begin{center}
\caption{Sample Tagged Sentences}
\end{center}
\end{table}
\vspace{-2.2cm}
\begin{figure}
\centering
\centerline{\includegraphics [scale=0.5] {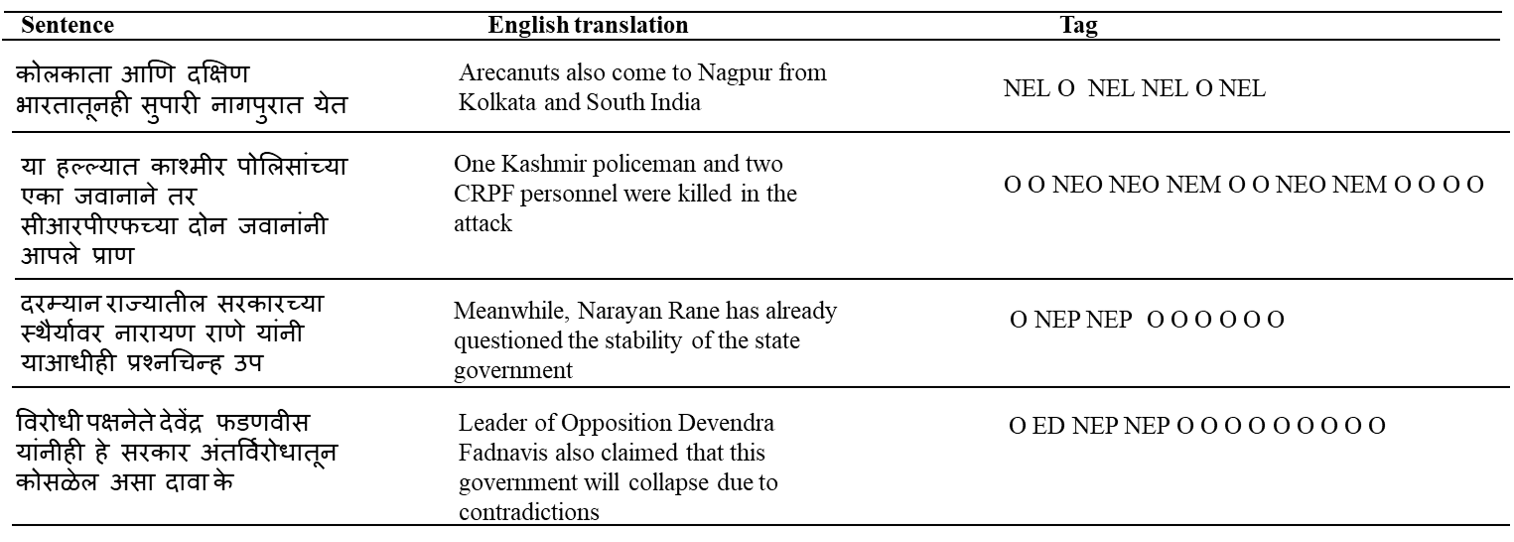}}
\end{figure}

\section{Conclusion}
In this paper, we hold forth on the problem of scarcity of annotated corpora and hence present L3Cube-MahaNER which is by far the largest dataset for Marathi Named Entity recognition, containing 25000 distinct sentences. We achieved the results using IOB and non-IOB notations on deep learning models such as CNN, LSTM, biLSTM, and transformers in BERT as listed above,  to set the basis for future work. We observed the highest accuracy on MahaRoBERTa for IOB notations and model MahaBERT for non-IOB notations. We believe that our corpus will play a pivotal role in expanding conversational AI for the Marathi Language.

\section*{Acknowledgements} This work was done under the L3Cube Pune mentorship program. We would like to express our gratitude towards our mentors at L3Cube for their continuous support and encouragement.

\bibliographystyle{splncs04}
\bibliography{main}
\end{document}